\documentclass[letterpaper]{article} 
\usepackage{aaai2026}  
\usepackage{times}  
\usepackage{helvet}  
\usepackage{courier}  
\usepackage[hyphens]{url}  
\usepackage{graphicx} 
\usepackage{natbib}  
\usepackage{caption} 
\usepackage{algorithm}
\usepackage{algorithmic}

\usepackage{amsmath}
\usepackage{ragged2e} 
\usepackage{booktabs,makecell, multirow, tabularx}
\usepackage{bbding}
\usepackage{color}
\usepackage{float}
\usepackage{amssymb}

\usepackage{newfloat}
\usepackage{listings}
\DeclareCaptionStyle{ruled}{labelfont=normalfont,labelsep=colon,strut=off} 
\floatstyle{ruled}
\newfloat{listing}{tb}{lst}{}
\floatname{listing}{Listing}
\title{MFmamba: A Multi-function Network for Panchromatic Image Resolution
Restoration Based on State-Space Model}
\author{
    Qian Jiang\textsuperscript{\rm 1},
    Qianqian Wang\textsuperscript{\rm 1},
    Xin Jin\textsuperscript{\rm 1}\equalcontrib\\
    Michal Wozniak\textsuperscript{\rm 2},
    Shaowen Yao\textsuperscript{\rm 1},
    Wei Zhou\textsuperscript{\rm 3}\equalcontrib\\
}
\affiliations{
    \textsuperscript{\rm 1}School of software, Yunnan University, Kunming, Yunnan, China\\
    \textsuperscript{\rm 2}Faculty of Information and Communication Technology, Wroclaw University of Science and Technology, Wroclaw, Poland\\
    \textsuperscript{\rm 3}School of Engineering, Yunnan University, State Key Laboratory of Vegetation Structure, Function and Construction (VegLab), Kunming, Yunnan, China\\

    jiangqian\_1221@163.com, qianqianwang1325@163.com, xinxin\_jin@163.com,  michal.wozniak@pwr.edu.pl, yaosw@ynu.edu.cn, zwei@ynu.edu.cn
}

\usepackage{bibentry}

\begin{document}

\maketitle

\begin{abstract}
Remote sensing images are becoming increasingly widespread in military, earth resource exploration. Because of the limitation of a single sensor, we can obtain high spatial resolution grayscale panchromatic (PAN) images and low spatial resolution color multispectral (MS) images. Therefore, an important issue is to obtain a color image with high spatial resolution when there is only a PAN image at the input. The existing methods improve spatial resolution using super-resolution (SR) technology and spectral recovery using colorization technology. However, the SR technique cannot improve the spectral resolution, and the colorization technique cannot improve the spatial resolution. Moreover, the pansharpening method needs two registered inputs and can not achieve SR. As a result, an integrated approach is expected. To solve the above problems, we designed a novel multi-function model (MFmamba) to realize the tasks of SR, spectral recovery, joint SR and spectral recovery through three different inputs. Firstly, MFmamba utilizes UNet++ as the backbone, and a Mamba Upsample Block (MUB) is combined with UNet++. Secondly, a Dual Pool Attention (DPA) is designed to replace the skip connection in UNet++. Finally, a Multi-scale Hybrid Cross Block (MHCB) is proposed for initial feature extraction. Many experiments show that MFmamba is competitive in evaluation metrics and visual results and performs well in the three tasks when only the input PAN image is used.
\end{abstract}

\begin{links}
    \link{Code}{https://github.com/QianqianWang1325/MFmamba.git}
\end{links}

\section{Introduction}
Remote sensing images are widely used in object detection \cite{10286153}, urban planning \cite{park2025efficient}, environmental monitoring \cite{10526301}, and other fields. They play an increasingly important role. PAN and MS images are different types of remote sensing images with information from different wavelengths of the same area \cite{feng2022deep}. PAN image has a high spatial resolution. Still, due to them represented as grayscale, the spectral resolution is low \cite{jin2024restoration}. The spectral resolution of MS image is higher, but due to physical limitations, the spatial resolution is lower. A single sensor cannot simultaneously capture remote sensing images with high spatial and spectral resolution. However, high-resolution (HR) color remote sensing images are more beneficial for interpreting remote sensing scenes, such as image fusion \cite{li2022deep}, semantic segmentation \cite{10192914}. Therefore, improving the spatial and spectral resolution of remote sensing images becomes a worthwhile research topic in the case of input-only PAN image.

The existing research uses SR technology to improve the  spatial resolution of low-resolution (LR) images and colorization technology to recover the spectral resolution of grayscale images. Therefore, the spatial resolution of PAN images can be restored using SR technology. Since the potential of convolutional neural networks (CNN) in image processing has been discovered, subsequent studies have introduced more CNN-based SR methods, such as \cite{li2023x}.  There are also many other methods to optimize image SR performance by different model structure, such as LGC-GDAN \cite{10403859}  was based on a generative adversarial network (GAN), MAT \cite{10935664} was based on Transformer, StableSR \cite{wang2024exploiting} was based on diffusion model. Image colorization is adopted to recover PAN image spectral resolution.  The automatic image colorization models based on deep learning (DL) have been gradually designed. For instance, CycleGAN-Color \cite{huang2021fully} was based on GAN,  \cite{kumar2021colorization} was based on Transformer, and Control\_Color \cite{liang2024control} was based on the diffusion model.

The purpose of PAN image joint SR and colorization is to promote the image's spatial resolution and spectral resolution synchronously. In existing methods, images SR and colorization were two separate goals. We designed a multi-function network (MFmamba) to solve these problems. Using different inputs, it can achieve three tasks, including PAN image SR, spectral recovery, joint SR and spectral recovery. The main contributions of our work are as follows: 
\begin{itemize}
	\item We design an efficient PAN image resolution restoration network (MFmamba) to produce colorized HR images so that MFmamba could realize the joint SR and spectral recovery. The proposed Mamba Upsample Block (MUB) adopts a state space model for resolution restoration. 
 
	\item We design a Multi-scale Hybrid Cross Block (MHCB) for shallow feature extraction, which can detect local and multi-scale features effectively and improve the ability of detail feature extraction.
	
	\item We introduce a novel Dual Pool Attention (DPA) to improve feature representation by dynamically adjusting channel weights so that the model may focus on more important feature channels.
\end{itemize}

\section{Related Work}
\subsection{Image Super-Resolution (SR)}
In recent years, image SR has made a breakthrough due to the strong learning capabilities of DL. Li \textit{et al}. \cite{li2020mdcn} designed an SR model, MDCN, which utilized dense connection and residual strategy to improve performance and used fewer parameters. Li \textit{et al}. \cite{li2023multi} proposed a multiscale factor-based hyperspectral image joint learning algorithm (MulSR), which used a symmetric guidance encoder and direction-sensing spatial context aggregation module to realize multiscale information interaction. Li \textit{et al}.\cite{li2025enhanced} released the power of deep image prior to the hyperspectral image SR and achieved high-quality reconstruction results. Zhu \textit{et al}. \cite{ZHU2023151} enhanced GAN using  residual blocks in the generator and attention mechanisms for feature fusion. It introduces RaGAN in the discriminator to improve edge and texture learning. Chen \textit{et al}. \cite{chen2023activating} reported a novel Hybrid Attention Transformer (HAT). They were  inspired by Swin Transformer and proposed a window-based self-attention, an overlapping cross-attention module for better cross-window interaction. Xie \textit{et al}. \cite{10935664} proposed a Multi-Range Attention Transformer (MAT) for SR tasks, which harnesses the computational efficiency of dilation operations and integrates multi-range attention (MA) with sparse multi-range attention (SMA) through self-attention mechanisms to capture features. 

Li \textit{et al}. \cite{li2022srdiff} proposed the first diffusion-based model (SRDiff) for single-image SR, which transforms Gaussian noise into SR images through a Markov chain and utilizes residual prediction to accelerate convergence. Wang \textit{et al}. \cite{wang2024exploiting} introduced a controllable feature wrapping module (StableSR) for balancing quality and fidelity via a scalar value and a progressive aggregation sampling strategy to adapt pre-trained diffusion models to any resolution. Guo \textit{et al}. \cite{guo2024mambair} proposed MambaIR, which introduced Residual State Space Block and channel attention mechanisms to enhance the ability of feature reconstruction. However, the above methods still have some problems. For example, extracting and retaining all textures is difficult, so error texture information will be generated in some places with complex details, and the detailed features extracted are insufficient.

\vspace{-0.3cm}
\subsection{Image colorization}
Traditional image colorization methods rely on user guidance. Automatic colorization methods are gradually designed with the development of DL. Feng \textit{et al}. \cite{feng2021remote} proposed an end-to-end CNN network for remote sensing image colorization, using multi-scale residual receptive fields for feature extraction and adopting residual, attention, and pixel-shuffle blocks. Saeed \textit{et al}. \cite{ANWAR2025102720} reviewed the image colorization techniques based on DL, pointed out the limitations of existing datasets, and introduced a new dataset for image colorization. Shafiq  \textit{et al}. \cite{shafiq2024transforming} designed an image colorization network using a hybrid architecture based on Transformer and GANs. They improved the encoder to generate color features and then used self-attention to capture long-range dependencies. GAN is also employed in image colorization tasks. Wu \textit{et al}. \cite{wu2021remote} proposed a new model for PAN image colorization by utilizing DCGAN with an auto-encoder-based generator and a residual network discriminator. Zhao \textit{et al}. \cite{9257445} proposed SCGAN. This fully automatic framework uses saliency maps and two hierarchical discriminators for perceptually accurate image colorization with minimal data, leveraging global features from a pre-trained VGG-16-Gray network. 

Jiang \textit{et al}. \cite{JIANG2025127091} designed a multiscale colorization network using a hybrid architecture based on Transformer and CNN. They used the attention and mask mechanism to enhance image features. Liang \textit{et al}. \cite{liang2024control}  developed a framework called CtrlColor, which leverages the pre-trained Stable Diffusion model to handle both unconditional and conditional colorization. The CtrlColor approach includes methods for encoding user-provided strokes, controlling color distribution and so on. However, these methods have some shortcomings. For example,  it cannot address color distortion problems and depends on the color distribution, making color results unreasonable\cite{9257445}. Therefore, grayscale image colorization methods still need further exploration.

\begin{figure*}[htp]
    \centering
    \centerline{\includegraphics[width=2\columnwidth]{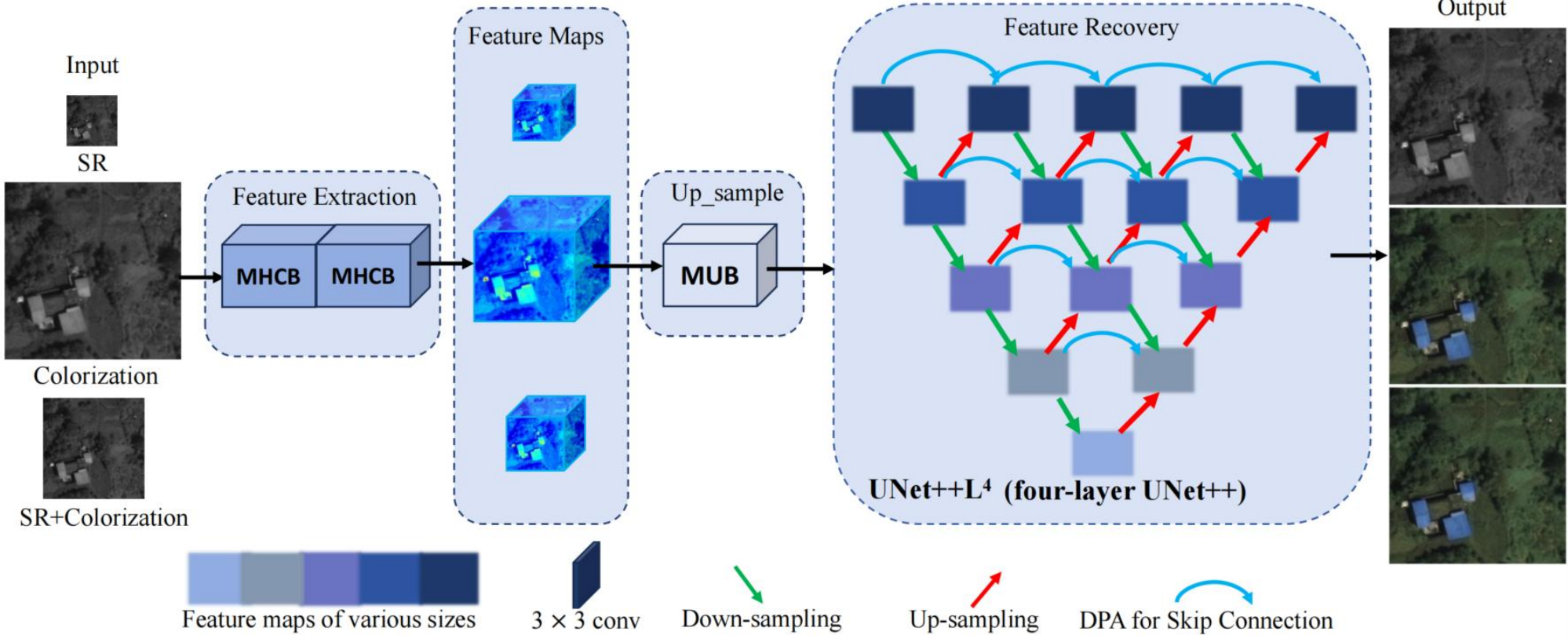}}
    \caption{General structure of the proposed solution (MFmamba).}
    \label{MFmamba}
    \vspace{-0.25cm}
\end{figure*}

\section{Proposed Method}
To achieve SR and spectral recovery tasks of PAN images, we propose a multi-function model (MFmamba) based on UNet++ \cite{zhou2018unet++} and mamba, which can realize the tasks of SR, spectral recovery, and joint SR and spectral recovery of PAN image. Fig. \ref{MFmamba} is the overall structure of our proposition. We combine the Mamba Upsample Module (MUB) constructed utilizing the state space model with UNet++ \cite{zhou2018unet++} and design a novel Dual Pool Attention (DPA) to replace the original skip connection in UNet++ for information transfer between the same-level feature maps. Multi-scale Hybrid Cross Block (MHCB) is designed to extract multi-scale features through convolution operations with different convolution kernel sizes.

\begin{figure}
    \centerline{\includegraphics[width=21pc]{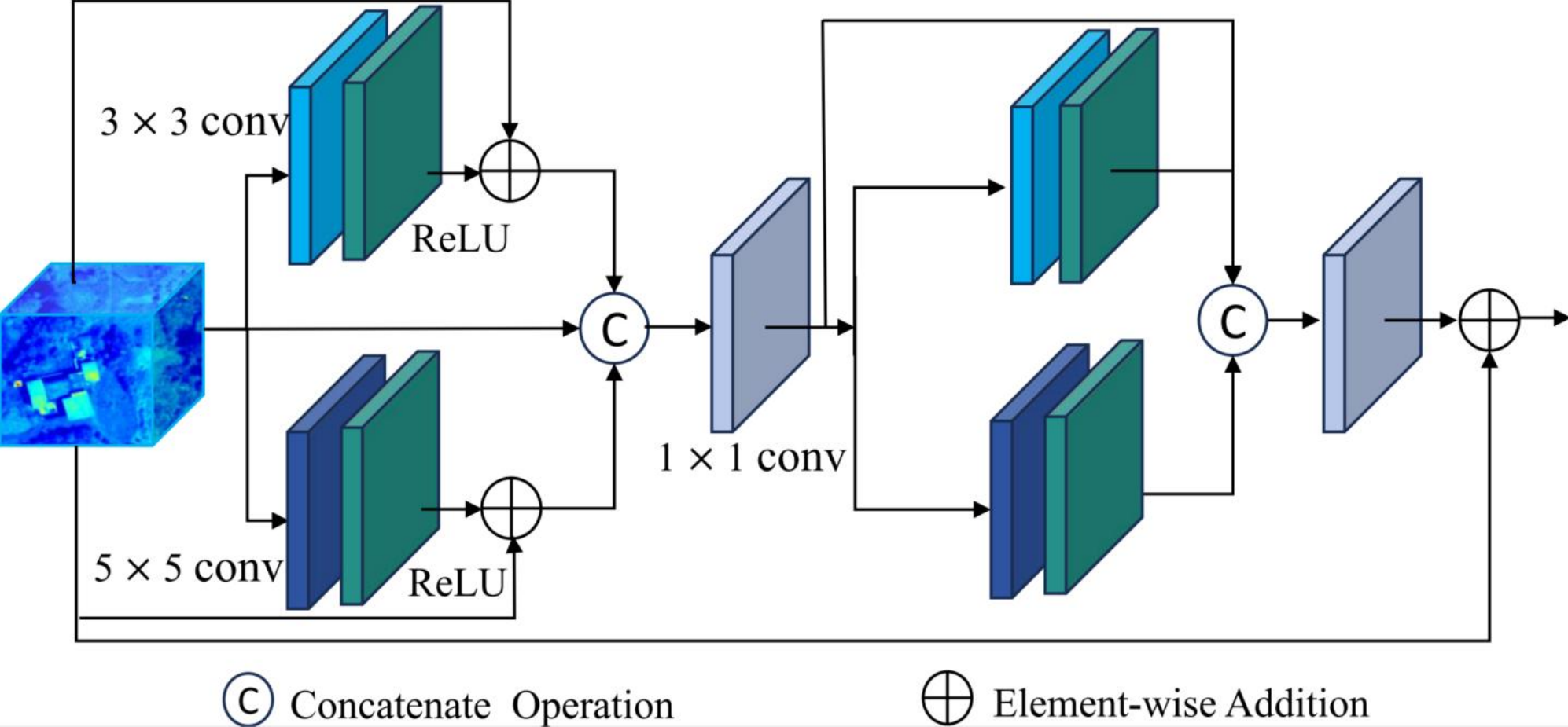}}
    \caption{The structure of Multi-scale Hybrid Cross Block (MHCB).}
    \label{MHCB}
    \vspace{-0.45cm}
\end{figure}

\subsection{Multi-scale Hybrid Cross Block (MHCB)}
We design a Multi-scale Hybrid Cross Block (MHCB) to extract multi-scale features information through parallel convolutional pathways with different kernel sizes, then skip connections with dense residual grouping to maintain gradient flow and reinforce critical feature persistence throughout the network hierarchy. Fig.~\ref{MHCB} shows the MHCB concept. We use $ 5 \times 5$ focuses more on global feature extraction than a small convolution kernel. We use $ 3 \times 3$ kernel with fewer parameters to obtain local features and fully utilize the detailed information in the PAN image. We also employ $ 1 \times 1$ kernel at the module's end to fuse features and combine the features of each layer. MHCB can be represented as follows:

\begin{equation}
\left\{
\begin{array}{l}
\scalebox{0.95}{$
\begin{aligned}
    X_{1} = ReLU(3 \times 3 Conv(X)) \oplus X, \\
    X_{2} = ReLU(5 \times 5 Conv(X)) \oplus X,
\end{aligned}
    $}
\end{array}
\right.
\end{equation}
\begin{equation}
    X_{3} = 1 \times 1 Conv(Concat(X_{1}, X_{2}, X)),
\end{equation}
\begin{equation}
\left\{
\begin{array}{l}
    X_{4} = ReLU(3 \times 3 Conv(X_{3})), \\
    X_{5} = ReLU(5 \times 5 Conv(X_{3})),
\end{array}
\right.
\end{equation}

\vspace{-0.3cm}
\begin{equation}
    MHCB_{out} = 1 \times 1 Conv((Concat(X_{3}, X_{4}, X_{5})) \oplus X,
\end{equation} 
where X represents the input feature of MHCB. $X_{1}$, $X_{2}$, $X_{3}$, $X_{4}$, and $X_{5}$ denote the intermediate output features. And $Concat$ represents the concatenate operation.

\subsection{Dual Pool Attention (DPA)}
\begin{figure}
    \centerline{\includegraphics[width=21pc]{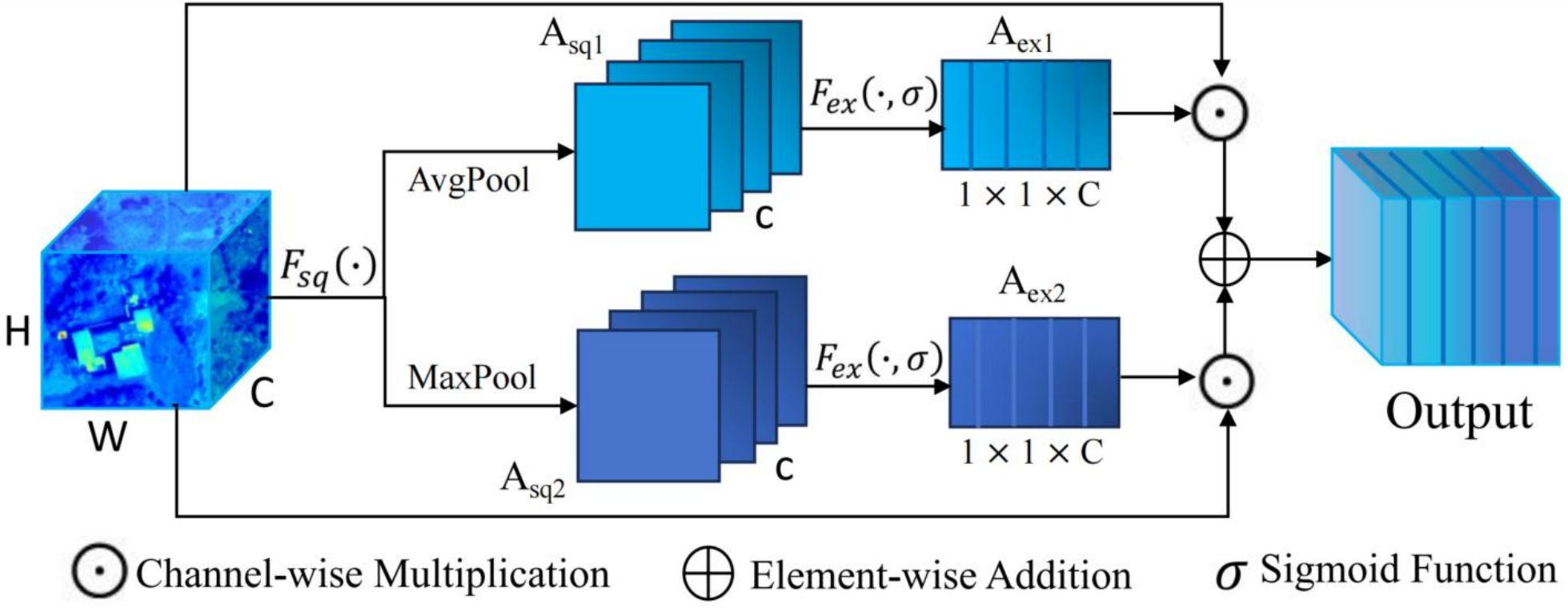}}
    \caption{The structure of Dual Pool Attention (DPA).}
    \label{DPA}
    \vspace{-0.45cm}
\end{figure}

In SR and the spectral recovery of PAN images, the attention mechanism is crucial in helping the model capture correlation and semantic information effectively. Therefore, we propose a novel Dual Pool Attention (DPA) as Fig.~\ref{DPA}. DPA employs a dual-stream architecture for comprehensive channel-wise feature calibration. DPA utilizes both adaptive global average pooling
(Eq. \ref{pool}) and maximum pooling to compress global spatial information into channel descriptors. Then, the sigmoid function is used to fully obtain the channel-wise dependencies and generate weight feature maps of size $1 \times 1 \times C$, see Eq. \ref{sigmoid}. These maps are applied to the input to evaluate channel importance and then summed (Eq. \ref{dpa}).  DPA introduces maximum pooling to capture significant features, enhancing model performance by focusing on fine details and important information for more comprehensive feature extraction. Combining the two pooling operations helps the model focus on key details and relevant information, helping the model better retain the features of the input image.

\vspace{-0.2cm}
\begin{equation}
\left\{
\begin{array}{l}
\scalebox{0.95}{$
\begin{aligned}

    A_{sq1} = AP_c(i, j) = \frac{1}{H \times W} \sum_{i=1}^{H} \sum_{j=1}^{W} X_{i,j,c}, \\
A_{sq2} = MP_c(i, j) = \max_{p, q \in \{1, \dots, k\}} X_{p,q,c},
\end{aligned}
\label{pool}
$}
\end{array}
\right.
\end{equation}

\begin{equation}
\left\{
\begin{array}{l}
    A_{ex1} =  sigmoid(A_{sq1}), \\
    A_{ex2} = sigmoid(A_{sq2}),
\label{sigmoid}
\end{array}
\right.
\vspace{-0.1cm}
\end{equation}

\begin{equation}
    DPA_{out} =  (X \odot A_{ex1}) \oplus (X \odot A_{ex1}), 
\label{dpa}
\end{equation}
where $X_{i,j,c}$ is the value at the feature map position $(i, j)$ on the channel $c$. $MP_c(i, j)$ indicates the value of position $(i, j)$ on channel $c$ of the pooled output feature map. $X_{p,q,c}$ is the pixel value in the pooled window $k \times k$.

\subsection{Mamba Upsampling Block (MUB)}
The structured state-space sequence model has made a breakthrough in recently, which maps a 1D sequence  $x(t) \in \mathbb{R} \rightarrow y(t) \in \mathbb{R}$ through an latent state  $h(t) \in \mathbb{R}^M$. Based on this, mamba model was proposed to solve the problem of low computational efficiency of Transformers on long sequences.

Specifically, the state space sequence model is defined with five parameters $(\Delta, \mathbf{A}, \mathbf{B}, \mathbf{C}, \mathbf{D})$, which sequence-to-sequence transformation is defined as a two-stage process. Formally, this system is represented as a linear ordinary differential equation (ODE)~\cite{guo2024mambair}: 
\begin{equation}
\begin{aligned}
    h'(t) &= \mathbf{A}h(t) + \mathbf{B}x(t), \quad
    y(t) = \mathbf{C}h(t) + \mathbf{D}x(t),
    \label{latent}
\end{aligned}
\end{equation}
where \emph{M} represents the state size, $\mathbf{A} \in \mathbb{R}^{M \times M}$, $\mathbf{B} \in \mathbb{R}^{M \times 1}$, $\mathbf{C} \in \mathbb{R}^{1 \times M}$, and $\mathbf{D} \in \mathbb{R}$.

The discretization process is then used to integrate Eq. \ref{latent} into an actual DL model. The parameter $\Delta$ controls the timescale, converting the continuous parameters $\mathbf{A}$ and $\mathbf{B}$ into their discrete counterparts, $\mathbf{\bar{A}}$  and $\mathbf{\bar{B}}$. The widely used  discretization approach is the zero-order hold (ZOH) defined as follows:
\begin{equation}
\begin{aligned}
    \mathbf{\bar{A}} = \exp(\Delta \mathbf{A}), \quad
    \mathbf{\bar{B}} = (\Delta \mathbf{A})^{-1}(\exp(\mathbf{A}) - \mathbf{I}) \cdot \Delta \mathbf{B}.
\end{aligned}
\end{equation}

After discretized with a time step size of $\Delta$, Eq. \ref{latent} can be reformulated as follows:
\vspace{-0.2cm}
\begin{equation}
\begin{aligned}
    h_\tau = \mathbf{\bar{A}} h_{\tau-1} + \mathbf{\bar{B}} x_\tau, \quad
    y_\tau = \mathbf{C}h_\tau + \mathbf{D}x_\tau,
    \label{ssm}
\end{aligned}
\vspace{-0.2cm}
\end{equation}
where $\tau$ represents a discrete time step, the original continuous time step $t$ is converted to the discrete time step $\tau$.
\begin{figure}
\vspace{-0.3cm}
    \centerline{\includegraphics[width=21pc]{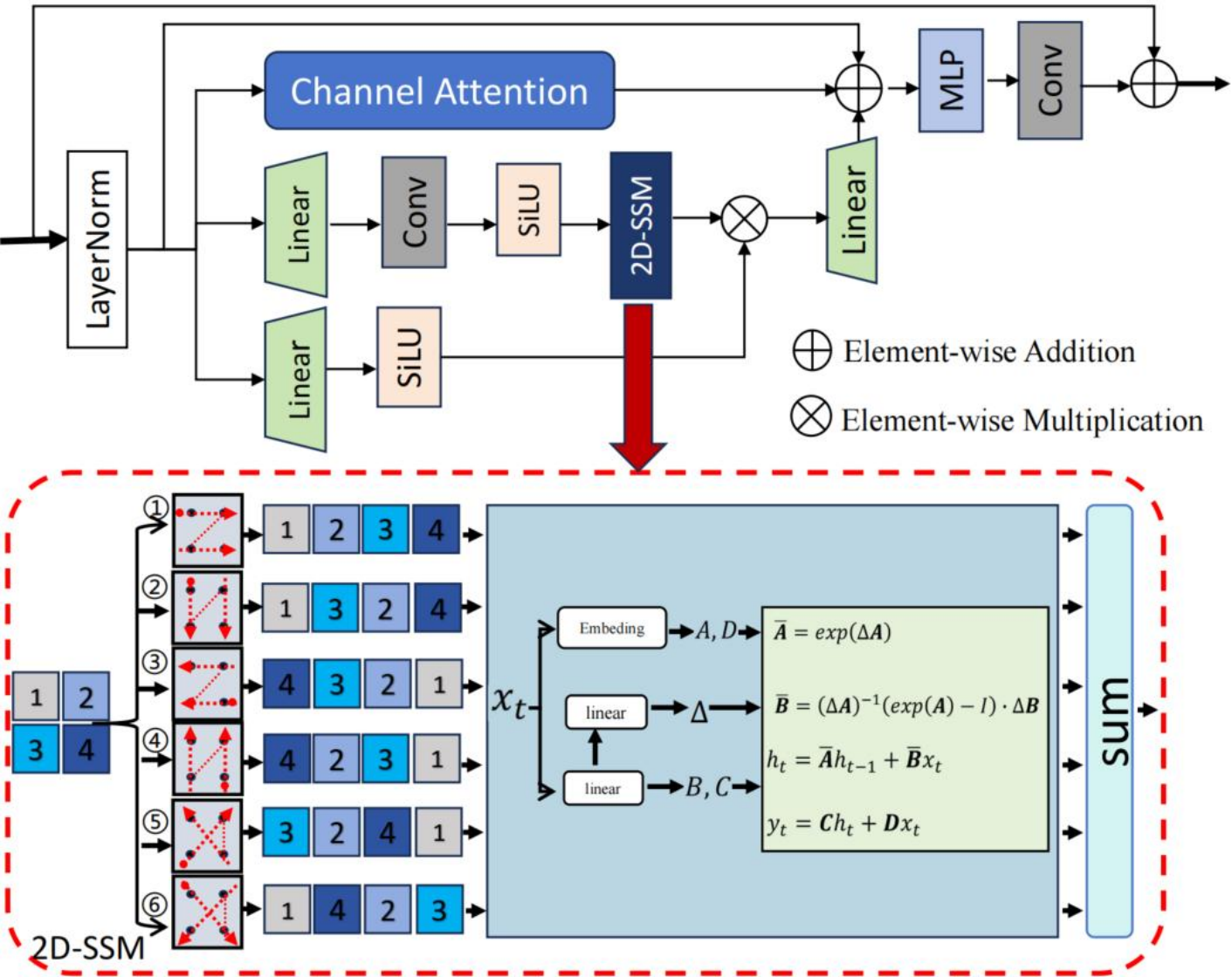}}
    \caption{The structure of Mamba Upsample Module (MUB).}
    \label{MUB}
    \vspace{-0.4cm}
\end{figure}

Eq. \ref{ssm} describes a linear time-invariant (LTI) system that remains static with different input parameters, and the system's equation of state never changes with the input. However, this static system has certain limitations. To overcome this problem, recent studies \cite{vmamba2024} proposed a selective scan mechanism that dynamically adjusts matrices $\mathbf{B}$, $\mathbf{C}$, and $\Delta$ based on input data, improving performance. Inspired by the good performance of this selective scan mechanism, we employ a 2D selective scan mechanism (2D-SSM) \cite{vmamba2024} to promote the awareness of contextual information embedded in the input \cite{guo2024mambair}. Finally, based on the above theory, we designed the Mamba Upsample Module (MUB). The MUB structure diagram is shown in Fig. \ref{MUB}. We increase the origina 2D image feature scanning from four different horizontal directions to six scanning directions, adding two diagonal scanning directions to capture data information more comprehensively. Then, we integrated and flattened the 2D feature maps into 1D sequences, and the long-range dependency of each sequence is computed by Eq. \ref{ssm}. Finally, the results are combined into a 2D feature map as the final result of MUB moudle \cite{guo2024mambair}. As shown the red box in Fig. \ref{MUB}. By utilizing complementary 1D traversal paths, the 2D-SSM framework allows each pixel to efficiently collect information from other pixels in multiple directions. Thus, it is convenient to establish the global receptive fields in 2D space \cite{vmamba2024}.

\vspace{-0.3cm}
\subsection{Loss Function}
We optimize our network using $L_1$ loss, which calculates the sum of absolute differences between the predicted and ground truth images as follows:
\vspace{-0.3cm}
\begin{equation}
    L_1(y, \hat{y}) = \sum_{i=1}^{n} |y_i - \hat{y}_i|,
    \label{L1loss}
    \vspace{-0.2cm}
\end{equation}
where $n$ denotes the number of batch training samples, $y_i$ represents the corresponding label image, and $ \hat{y}_i$ stands for the output image generated by MFmamba.
\begin{table}
	\begin{center}
		\caption{Ablation experiments  on  Potsdam, black bold font marks the best performance.} 
		\label{ablation}
		\renewcommand\arraystretch{1.5}
		\setlength{\tabcolsep}{2mm}{
                \resizebox{\linewidth}{14mm}{
			\begin{tabular}{cccccccc}
				\hline
				   & MHCB & DPA & MUB & PSNR$\uparrow$ & SSIM$\uparrow$ & MSE$\downarrow$ & MAE$\downarrow$ \\
				\hline
                    w/o DPA & \Checkmark & \XSolidBrush & \Checkmark &39.980&0.965&7.457&76.791\\
                    w/o MUB & \Checkmark & \Checkmark & \XSolidBrush &39.794&0.965&7.679&\textbf{68.398}\\
                    w/o MHCB & \XSolidBrush & \Checkmark & \Checkmark &40.061&0.966&7.202&71.327\\
                    w/ MHCB-1 & \Checkmark & \Checkmark & \Checkmark &40.072&0.966&7.210&75.845\\
                    w/ MHCB-2(Ours) & \Checkmark & \Checkmark & \Checkmark &\textbf{40.148}&\textbf{0.967}&\textbf{7.096}&73.499\\
                    w/ MHCB-3 & \Checkmark & \Checkmark & \Checkmark &40.080&0.966&7.204&72.915\\
				\hline 
		\end{tabular}}}
	\end{center}
 \vspace{-0.4cm}
\end{table}

\section{EXPERIMENTS AND ANALYSIS}
\begin{table}
\vspace{-0.1cm}
	\begin{center}
		\caption{Different UNet++ depth on  Potsdam.} 
		\label{depth}
		\renewcommand\arraystretch{1.5}
		\setlength{\tabcolsep}{2mm}{
                \resizebox{\linewidth}{8mm}{
			\begin{tabular}{ccccccc}
				\hline
				   UNet++ depth& PSNR$\uparrow$ & SSIM$\uparrow$ & MSE$\downarrow$ & MAE$\downarrow$ & SAM$\downarrow$& LPIPS$\downarrow$\\
				\hline
                    2 &39.548&0.965&7.926&77.036&0.039&0.038\\
                    3 &39.986&0.965&7.382&75.545&0.030&0.039\\
                    4(ours) &\textbf{40.148}&\textbf{0.967}&\textbf{7.096}&\textbf{73.499}&\textbf{0.029}&\textbf{0.036}\\
				\hline 
		\end{tabular}}}
	\end{center}
 \vspace{-0.6cm}
\end{table}

\begin{table}
	\begin{center}
		\caption{Different scanning methods and patch on  Potsdam.} 
		\label{scan}
		\renewcommand\arraystretch{1.5}
		\setlength{\tabcolsep}{2.0mm}{
                \resizebox{\linewidth}{12mm}{
			\begin{tabular}{cccccc}
				\hline
				   Configuration& PSNR$\uparrow$ & SSIM$\uparrow$ & MSE$\downarrow$ & MAE$\downarrow$ & SAM$\downarrow$\\
				\hline
                    $3\times3$ patch &39.266&0.966&7.828&75.391&0.039\\
                    w/o MUB &39.794&0.965&7.679&\textbf{68.398}&0.034\\
                    w/ MUB(\textcircled{\scriptsize 5}\_scan) &40.071&0.966&7.185&73.53&0.031\\
                    w/ MUB(\textcircled{\scriptsize 6}\_scan)&40.007&0.966&7.34&74.477&0.032\\
                    $2\times2$ MUB(ours) &\textbf{40.148}&\textbf{0.967}&\textbf{7.096}&73.499&\textbf{0.029}\\
				\hline 
		\end{tabular}}}
	\end{center}
    \vspace{-0.6cm}
\end{table}

\subsection{Dataset and Setting}
We evaluated MFmamba on five remote sensing datasets: NWPU, Potsdam, QuickBird, GF2, and RSSCN7. We chose MSE, PSNR, SSIM, MAE, SAM, and LPIPS as the model's performance metrics. The depth of $d$ UNet++ \cite{zhou2018unet++} for MFmamba was set at 4, and the growth rate of $g$ was 32. In all experiments, the batch size was 1, and the number of training epochs was 32. The random seed was set to 10. We optimized the proposed MFmamba model using the Adam optimizer with a learning rate of  $1 \times 10^{-4}$, and adjusted the learning rate every 10 iterations using StepLR. Moreover, ${\beta}_1$ = 0.9 and ${\beta}_2$ = 0.99 were chosen. All experiments were based on  PyTorch, and all models were trained by a 80 GB memory GPU. For the SR x2 task, we downsampled the original PAN image to $128 \times 128$ as the input. For the SR x4, we downsampled the original PAN image to $64 \times 64$ as the input. In joint SR and colorization tasks, we downsampled the original PAN image to $128 \times 128$ as the input. 

\subsection{Ablation Study}
To verify the performance of the proposed modules (MHCB, DPA, and MUB), we designed an ablation experiment on the Potsdam dataset for the joint SR and spectral recovery task of remote sensing images. The SR scale factor was set to x2 and all experiments used the same configuration. MHCB, DPA, and MUB are independent of each other. First, we test the performance of the DPA module and MUB module. As seen in Table \ref{ablation}, without the DPA module and MUB module led to degraded performance. This demonstrates that the DPA and MUB module helps the model focus on critical information, preserves key image features, improves the upsampling process, and enhances overall performance. Second, we estimated the impact of the MHCB module. Our model employed two MHCBs for initial feature extraction. To determine the optimal number, we conducted an ablation study using 1, 2, and 3 MHCBs, finally finding that using two MHCBs brings the best performance, as shown in the third to sixth rows of Table \ref {ablation}.

We also evaluated the performance of different UNet++ \cite{zhou2018unet++} depths in this study. We test the commonly used two-layer, three-layer, and four-layer UNet++ architectures. Table \ref{depth} shows that the four-layer network we adopted demonstrated superior performance, excelling in feature extraction and recovery. In addition. We used inputs with 4 patches and 9 patches in MUB to verify the impact of different numbers of patches on the model's performance. We also conducted ablation experiments on the two scanning directions (referred to as \textcircled{\scriptsize 5} and \textcircled{\scriptsize 6}) designed by us in the 2D-SSM module of MUB, in addition to the traditional four horizontal scanning directions. All results are shown in the Table \ref{scan}. It can be observed that the 4-patch configuration used in our model achieves better performance than the 9-patch configuration. Moreover, the MUB model cannot achieve optimal performance when only the \textcircled{\scriptsize 5} or \textcircled{\scriptsize 6} scanning direction is used. The model’s performance is effectively improved when both scanning directions are incorporated, demonstrating the effectiveness of our proposed two scanning directions.

\subsection{Experimental Comparison and Discussions}
We performed three kinds of experiments to test the performance of our method.
\begin{table*}
\centering
\caption{Contrast experiments results on Potsdam and NWPU, black bold font marks the best performance.}
\label{potsdan-nwpu}
\renewcommand\arraystretch{1.5}
		\setlength{\tabcolsep}{1.5mm}{
            \resizebox{0.90\linewidth}{27mm}{
\begin{tabular}{l c c c c c c| c c c c c c c}
\hline
\multirow{2}{*}{Method} & \multicolumn{6}{c}{Potsdam Dataset} & \multicolumn{6}{c}{NWPU Dataset} \\
\cline{2-13}
  &PSNR$\uparrow$ & SSIM$\uparrow$ & MSE$\downarrow$ & MAE$\downarrow$ & SAM$\downarrow$& LPIPS$\downarrow$&PSNR$\uparrow$ & SSIM$\uparrow$ & MSE$\downarrow$ & MAE$\downarrow$ & SAM$\downarrow$& LPIPS$\downarrow$ \\
\hline
MBPRR\cite{jin2024restoration}&34.953&0.943&28.653&103.165&\textbf{0.023}&0.064&28.504&0.913&101.887&128.501&0.077&0.096\\
CASR\cite{liu2022casr}&32.685&0.946&51.054&105.339&0.029&0.079 &28.200&0.906&109.594&120.439&\textbf{0.049}&0.105\\
RSI\cite{feng2022deep}&34.857&0.910&25.760&99.911&0.054&0.052& 32.354&0.913&59.534&\textbf{104.111}&0.074&\textbf{0.068}\\
CSRDNN\cite{feng2021csrdnn}&31.461&0.954&56.084&96.923&0.081&0.049& 31.478&0.905&62.645&127.096&0.087&0.070\\
HAT\cite{chen2023activating}+CIR\cite{feng2021remote}&32.853&0.958&34.774&103.314&0.062&0.039&22.760&0.533&393.884&125.429&0.299&0.279\\
CIR\cite{feng2021remote}+HAT\cite{chen2023activating}&27.777&0.720&375.405&113.637&0.147&0.082&24.519&0.672&257.197&126.742&0.163&0.467\\
SwinIR\cite{liang2021swinir}+CIR\cite{feng2021remote}&32.838&0.957&34.907&103.308&0.062&0.038&22.789&0.536&390.958&125.484&0.198&0.270\\
CIR\cite{feng2021remote}+SwinIR\cite{liang2021swinir}&27.777&0.720&375.272&113.547&0.147&0.082&24.518&0.672&257.232&126.674&0.163&0.469\\
Ours & \textbf{40.148}&\textbf{0.966}&\textbf{7.096}&\textbf{73.499}&0.029&\textbf{0.036}&\textbf{33.183}&\textbf{0.927}&\textbf{46.606}&106.451&0.066&0.074\\
\hline
\end{tabular}}}

\end{table*}

\begin{table*}
\vspace{-0.5cm}
\centering
\caption{Experimental results of SR at different up\_scale factors on Potsdam, black bold font marks the best performance.}
\label{sr_difscale}
\renewcommand\arraystretch{1.5}
		\setlength{\tabcolsep}{1.5mm}{
            \resizebox{0.9\linewidth}{24mm}{
\begin{tabular}{l c c c c c c| c c c c c c c}
\hline
\multirow{2}{*}{Method} & \multicolumn{6}{c}{x2} & \multicolumn{6}{c}{x4} \\
\cline{2-13}
  &PSNR$\uparrow$ & SSIM$\uparrow$ & MSE$\downarrow$ & MAE$\downarrow$ & SAM$\downarrow$& LPIPS$\downarrow$&PSNR$\uparrow$ & SSIM$\uparrow$ & MSE$\downarrow$ & MAE$\downarrow$ & SAM$\downarrow$& LPIPS$\downarrow$ \\
\hline
Bicubic&35.591&0.908&21.307&112.893&0.049&0.130&31.145&0.788&59.929&117.396&0.082&0.410\\
RSI\cite{feng2022deep} &37.525&0.936&13.626&\textbf{66.786}&0.038&0.055 &31.886&\textbf{0.810}&49.420&\textbf{68.907}&0.072&0.189\\
MDCN\cite{li2020mdcn} &34.718&0.859&313.392&99.817&0.096&0.0706&26.187&0.668&314.211&101.733&0.091&0.071\\
HAT\cite{chen2023activating}&34.963&0.861&311.239&106.593&0.095&0.0705&30.209&0.752&328.867&113.026&0.121&0.073\\
SwinIR\cite{liang2021swinir}&34.921&0.861&311.239&106.423&0.095&0.0704&30.006&0.746&331.536&110.928&0.122&0.073\\
MulSR\cite{li2023multi} &25.593&0.903&343.201&144.138&0.089&0.075&23.975&0.885&393.033&102.323&0.130&0.116\\
Ours &\textbf{38.458}&\textbf{0.954}&\textbf{10.129}&77.288&\textbf{0.035}&\textbf{0.033}& \textbf{32.257}&0.809&\textbf{46.230}&95.163&\textbf{0.072}&\textbf{0.019}\\
\hline
\end{tabular}}}
\end{table*}

\begin{figure*}
    \centering
    \centerline{\includegraphics[width=2\columnwidth]{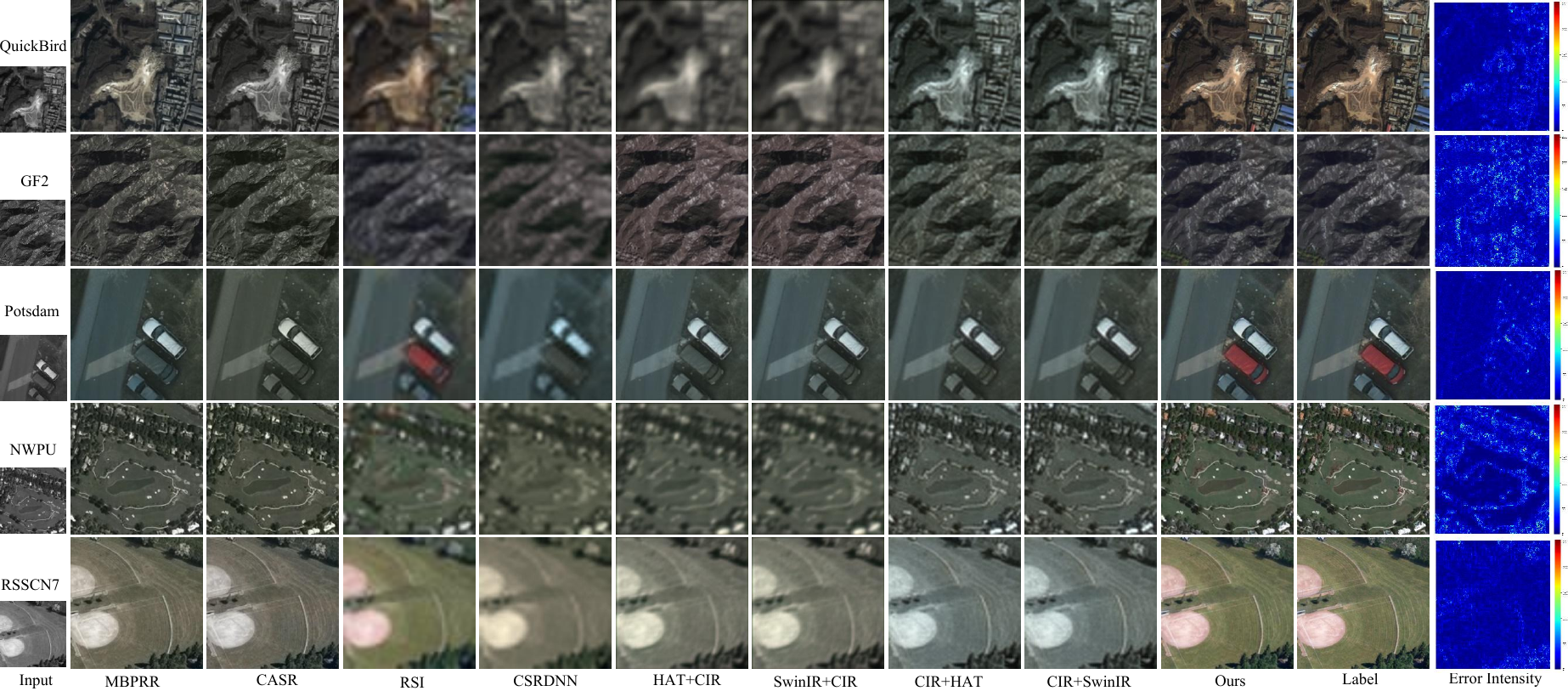}}
    \caption{Visual results of MFmamba and benchmark methods on five different datasets.}
    \label{results3}
\end{figure*}

\subsubsection{Joint SR and Spectral Recovery}
Our primary goal is to achieve joint PAN image SR and spectral recovery for remote sensing images. We compare our model with some existing models for joint SR and colorization, including RSI \cite{feng2022deep}, CSRDNN \cite{feng2021csrdnn}, CASR \cite{liu2022casr}, and MBPRR \cite{jin2024restoration}, using a x2 scale factor for the SR task. HAT \cite{chen2023activating} + CIR \cite{feng2021remote} means that HAT is used first for SR operation and then the CIR is used for colorization. CIR \cite{feng2021remote} + HAT \cite{chen2023activating} means using the CIR colorization model first and then HAT SR model. SwinIR \cite{liang2021swinir} + CIR \cite{feng2021remote} and CIR \cite{feng2021remote} + SwinIR \cite{liang2021swinir} have the same meaning as above. Comparative experiments and analyses were conducted on two datasets to thoroughly assess the generalization capabilities and performance of MFmamba across different datasets. The experimental results  on Potsdam and NWPU are presented in Tables \ref{potsdan-nwpu}, experimental results on QuickBird, GF2 and  RSSCN dataset in the Appendix for details. As shown in these tables, while MFmamba may be slightly inferior in a few metrics, it achieves the best overall performance. This demonstrates MFmamba's superiority in jointly accomplishing SR and spectral recovery tasks. It also indicates that MFmamba excels at extracting details and critical information in remote sensing images, enhancing feature recovery performance. 
\begin{figure}
\vspace{-0.3cm}
    \centerline{\includegraphics[width=14pc,height=10pc]{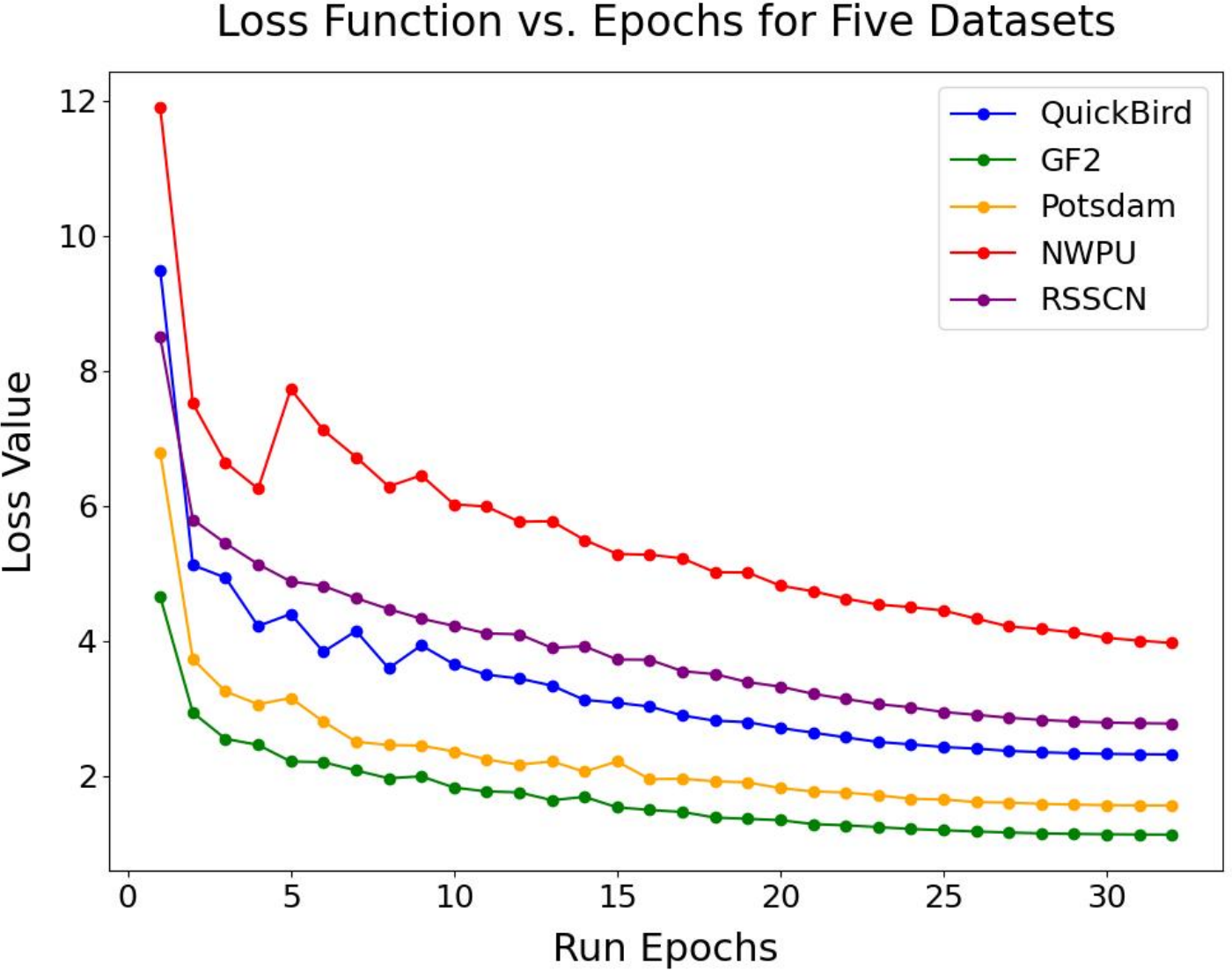}}
    \caption{The curves of the loss function with the number of iterations on the five datasets.}
    \label{5_loss}
    \vspace{-0.5cm}
\end{figure}

The convergence curves of the loss function values of the proposed MFmamba method with the number of iterations on five datasets as shown in Fig. \ref{5_loss}. To better illustrate our experimental results, we selected one image from each dataset to compare the performance of several existing methods with our proposed MFmamba, as shown in Fig. \ref{results3}. The 'Error Intensity' in the last column represents the error intensity between the generated results of our model and the label. The color changes gradually from blue to red. Among them, blue represents the area with relatively small errors, while yellow and red represent the areas with relatively large errors. 

\subsubsection{Super-Resolution}
\begin{figure}[h]
    \centering
    \centerline{\includegraphics[width=1\columnwidth]{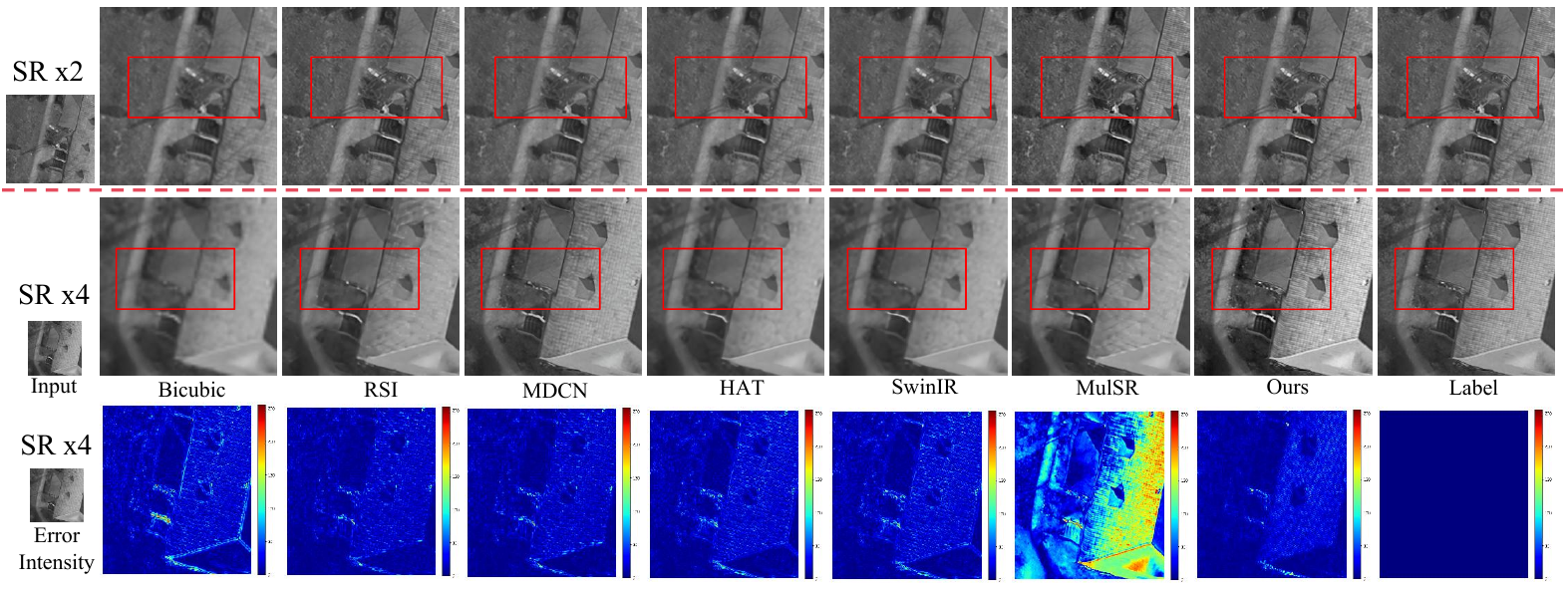}}
    \caption{Visual SR results of MFmamba and comparison methods on Potsdam dataset.}
    \label{SR_results}
    \vspace{-0.3cm}
\end{figure}

For the image SR task, we compared MFmamba with bicubic interpolation as the baseline and comparison methods including RSI \cite{feng2022deep}, MDCN \cite{li2020mdcn}, HAT \cite{chen2023activating}, SwinIR \cite{liang2021swinir}, and MulSR \cite{li2023multi}. We conducted comparative experiments and analyses exclusively on the Potsdam dataset. We performed experiments using x2 and x4 upscale factors for the SR task. The results are shown in Table \ref{sr_difscale}. To illustrate the result further, we select two images from the Potsdam dataset to present the results of the comparison methods and MFmamba on x2 and x4 upscale factors, as shown in Fig. \ref{SR_results}. These results demonstrate that MFmamba excels at jointly performing SR and spectral recovery and achieves outstanding performance when focused solely on SR tasks. MFmamba performs better on images SR of trees, roofs, buildings, etc. (see the red box in Fig. \ref{SR_results}). In Fig. \ref{SR_results}, the last row is the error intensity heatmap of the SR results generated by various comparative methods for the image of SRx4 and the Label image, which can intuitively display the error distribution and detail differences of different methods. 

\begin{table}
	\begin{center}
		\caption{Colorization experimental results
 on  Potsdam.}
		\label{color}
		\renewcommand\arraystretch{1.5}
		\setlength{\tabcolsep}{1.5mm}{
            \resizebox{\linewidth}{16mm}{
			\begin{tabular}{ccccccc}
				\hline
				   Method& PSNR$\uparrow$ & SSIM$\uparrow$ & MSE$\downarrow$ & MAE$\downarrow$ & SAM$\downarrow$&LPIPS$\downarrow$\\
				\hline
                RSI\cite{feng2022deep}&33.135&0.983&47.805&103.962&0.068& 0.052\\
                CIR\cite{feng2021remote} &31.990&0.957&42.428&95.134&0.069&0.077\\
                SEGAN\cite{wu2019remote} &32.513&0.632&396.840&122.054&0.083&0.640\\
                Wu\cite{wu2019remote} &31.558&0.537&458.715&127.489&0.085&0.634\\
                Iizukas\cite{iizuka2016let} &11.072&0.190&507.459&125.625&0.465&0.110 \\
                Huang\cite{huang2021fully} &32.025&0.968&72.866&105.335&0.066&0.175\\
                Ours &\textbf{35.569}&\textbf{0.984}&\textbf{28.477}&\textbf{88.036}&\textbf{0.052}&\textbf{0.048}\\
				\hline 
		\end{tabular}}}
	\end{center}
 \vspace{-0.6cm}
\end{table}

\subsubsection{Spectral Recovery}
\begin{figure}[!h]
    \centering
    \centerline{\includegraphics[width=1\columnwidth]{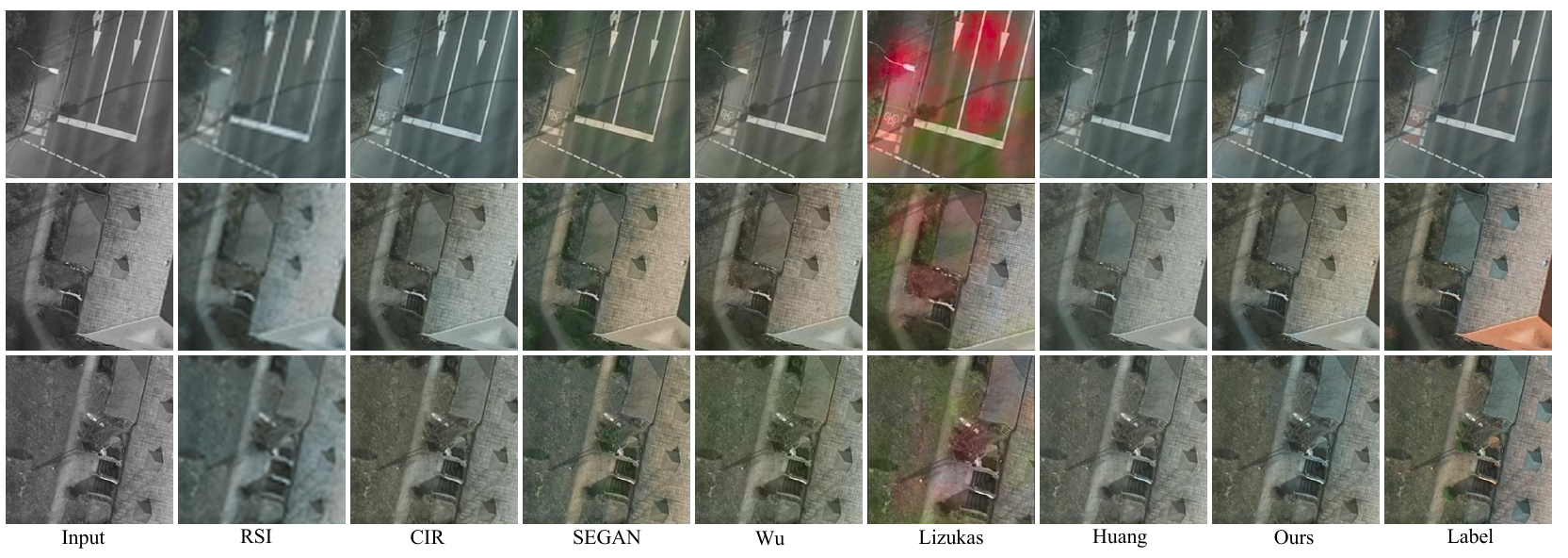}}
    \caption{Visual colorization results of MFmamba and benchmark methods on Potsdam dataset.}
    \label{color_results}
    \vspace{-0.5cm}
\end{figure}

The propose image colorization method was compared with the most advanceding algorithms include RSI \cite{feng2022deep}, CIR \cite{feng2021remote}, SEGAN \cite{wu2019remote}, Wu \cite{wu2019remote}, Iizukas \cite{iizuka2016let} and Huang \cite{huang2021fully} on the Potsdam dataset. The results are presented in Table \ref{color}. Although image colorization is an additional function of MFmamba model, it still performs better than other methods, and the generated results have more visual advantages. We selected five images from the Potsdam dataset to show the results of several comparison methods adopted and the results of MFmamba, as shown in Fig. \ref{color_results}. MFmamba performed the best in image colorization tasks, generating images with more realistic and natural colors, rich details, and very close to real labeled images. Both color transitions and edge details have been effectively restored, demonstrating excellent spectral recovery ability and outstanding visual effects.

\subsubsection{Comparison Between Mamba and Transformer}
\begin{figure}[H]
    \centerline{\includegraphics[width=20pc]{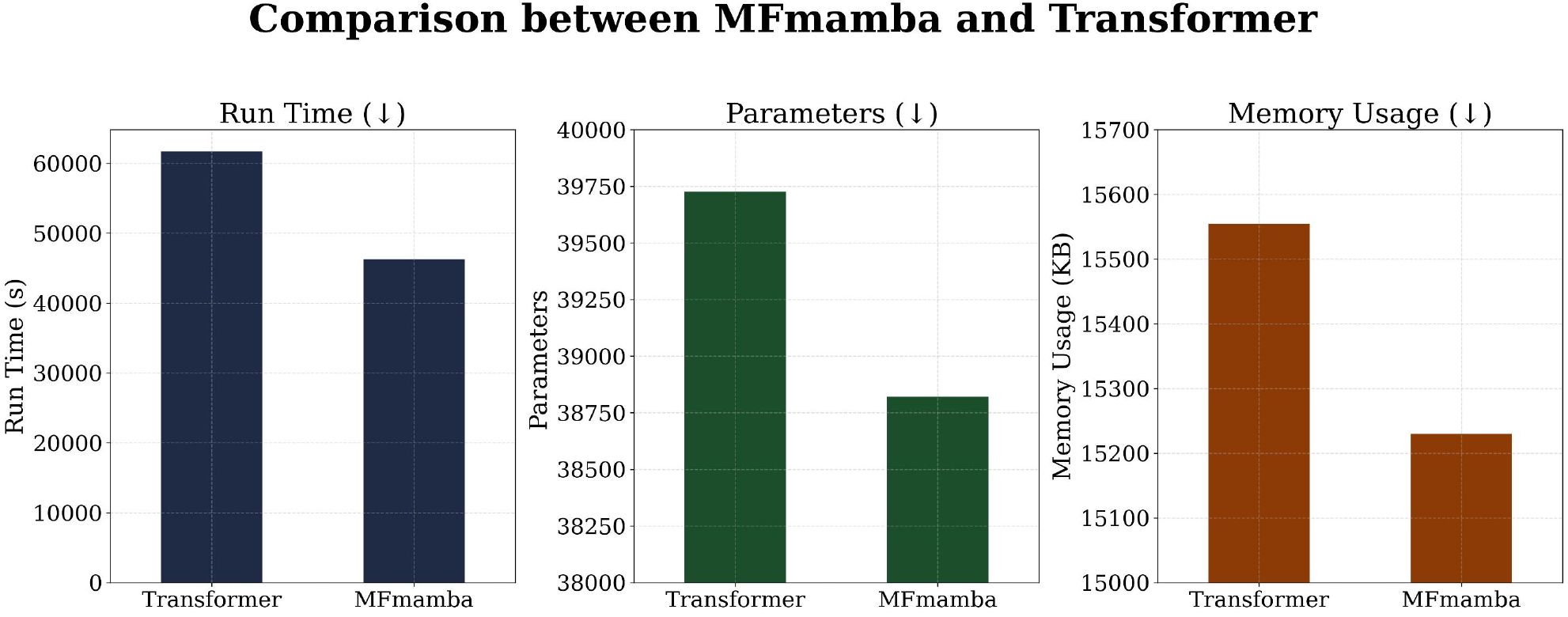}}
    \caption{The bar chart for comparison between Mamba and Transformer on Potsdam.}
    \label{trans_mamba}
    \vspace{-0.3cm}
\end{figure}

We conducted experiments by replacing the mamba module in the proposed MUB module with the Transform module to compare and analyse the differences between mamba and Transform in terms of running time, number of parameters and memory usage on the Potsdam dataset. The experimental results are shown in Fig. \ref{trans_mamba}, from which we can see that the mamba-based model we designed runs faster and has fewer parameters then Transformer, demonstrating the powerful advantages of Mamba. 

\section{Conclusion}
We proposed a multi-function resolution recovery model called MFmamba to enhance the spatial and spectral resolution of remote sensing images. MFmamba employs a state space model to develop the MUB module, which was used to enhance the model's contextual modeling capabilities. The proposed MHCB extracts multiscale features, effectively conveying detailed image information. We introduced a novel attention mechanism, DPA, to enable the model to focus on the most important feature channels. Extensive experiments and visualization results demonstrate that MFmamba, combined with these modules, achieves impressive results in spatial resolution recovery, spectral resolution recovery, and joint recovery in PAN images. Furthermore, our method outperforms most existing super-resolution and colorization models. Extensive experiments also validate the effectiveness of integrating multiple tasks within a single framework.

\section{Acknowledgments}
This study is supported by the National Natural Science Foundation of China (Nos. 62261060), Yunnan Fundamental Research Projects (Nos. 202503AG380006, 202301AW070007, 202301AU070210, 202401AT070470), and Yunnan Province Expert Workstations (202305AF150078), Yunnan Province Special Project (Grant No.202403AP140021), and Xingdian Talent Project in Yunnan Province of China.

\bibliography{aaai2026}

\end{document}